\documentclass{article}

     \PassOptionsToPackage{numbers, compress}{natbib}
 \usepackage[final]{neurips_2020}

\usepackage[utf8]{inputenc} % allow utf-8 input
\usepackage[T1]{fontenc}    % use 8-bit T1 fonts
\usepackage{hyperref}       % hyperlinks
\usepackage{url}            % simple URL typesetting
\usepackage{booktabs}       % professional-quality tables
\usepackage{amsfonts}       % blackboard math symbols
\usepackage{nicefrac}       % compact symbols for 1/2, etc.
\usepackage{microtype}      % microtypography
\usepackage{amsmath,amssymb,amsfonts}
\usepackage{algorithm}
\usepackage{algpseudocode}
\usepackage{graphicx}
\usepackage{textcomp}
\usepackage{xcolor}
\usepackage{stfloats}
\usepackage{subcaption}

\title{DBS: Dynamic Batch Size For Distributed Deep Neural Network Training}

\author{%
  Qing Ye, Yuhao Zhou, Mingjia Shi, Yanan Sun, Jiancheng Lv\\
  \texttt{yeqing@scu.edu.cn,sooptq@gmail.com,Jiancheng Lv@scu.edu.cn}\\
  \texttt{College of Computer Science, Sichuan University}
}

\begin{document}

\maketitle

\begin{abstract}
 Synchronous strategies with data parallelism, such as the Synchronous Stochastic Gradient Descent (S-SGD) and the model averaging methods, are widely utilized in distributed training of Deep Neural Networks (DNNs), largely owing to its easy implementation yet promising performance. Particularly, each worker of the cluster hosts a copy of the DNN and an evenly divided share of the dataset with the fixed mini-batch size, to keep the training of DNNs convergence. In the strategies, the workers with different computational capability, need to wait for each other because of the synchronization and delays in network transmission, which will inevitably result in the high-performance workers wasting computation. Consequently, the utilization of the cluster is relatively low. To alleviate this issue, we propose the Dynamic Batch Size (DBS) strategy for the distributed training of DNNs. Specifically, the performance of each worker is evaluated first based on the fact in the previous epoch, and then the batch size and dataset partition are dynamically adjusted in consideration of the current performance of the worker, thereby improving the utilization of the cluster. To verify the effectiveness of the proposed strategy, extensive experiments have been conducted, and the experimental results indicate that the proposed strategy can fully utilize the performance of the cluster, reduce the training time, and have good robustness with disturbance by irrelevant tasks. Furthermore, rigorous theoretical analysis has also been provided to prove the convergence of the proposed strategy. 
\end{abstract}

\section{Introduction}
Generally, the large-scale Deep Neural Network~(DNN) is shared by all computational workers in the distributed training of the data parallelism~\cite{Dean2012Large,elastic}. During the training, a worker computes local gradient on its sub-dataset after each iteration, and then all sub-gradients are accumulated essentially through well-design strategies to compute the new parameters of the DNN. Lastly, the new parameters are redistributed to all workers and the distributed training of the DNN continues. The main challenge comes with such circumstance is how to effectively and efficiently collect gradients among multiple computational workers. In practice, the synchronous strategies are widely employed to aggregate the sub-gradients, such as Bulk Synchronous Parallel (BSP)~\cite{bsp}, parallel Synchronous Stochastic Gradient Decent(S-SGD)~\cite{PSG,PRSGD}, to name a few. The characteristic of the synchronous methods is that the computational workers do not start the next iteration until they all commit their local gradients and receive the new global parameters. However, owing to the different performance of the workers, some workers may calculate the local gradients faster while others may be slower, which will result in the mutual waiting among the workers because of the essential synchronization. In other words, the faster worker would waste computational resource and thereby decreasing the utilization of the cluster~\cite{SYPS,COS}.
 
 \begin{figure}
        \begin{subfigure}{0.45\textwidth}
            \includegraphics[width=\linewidth]{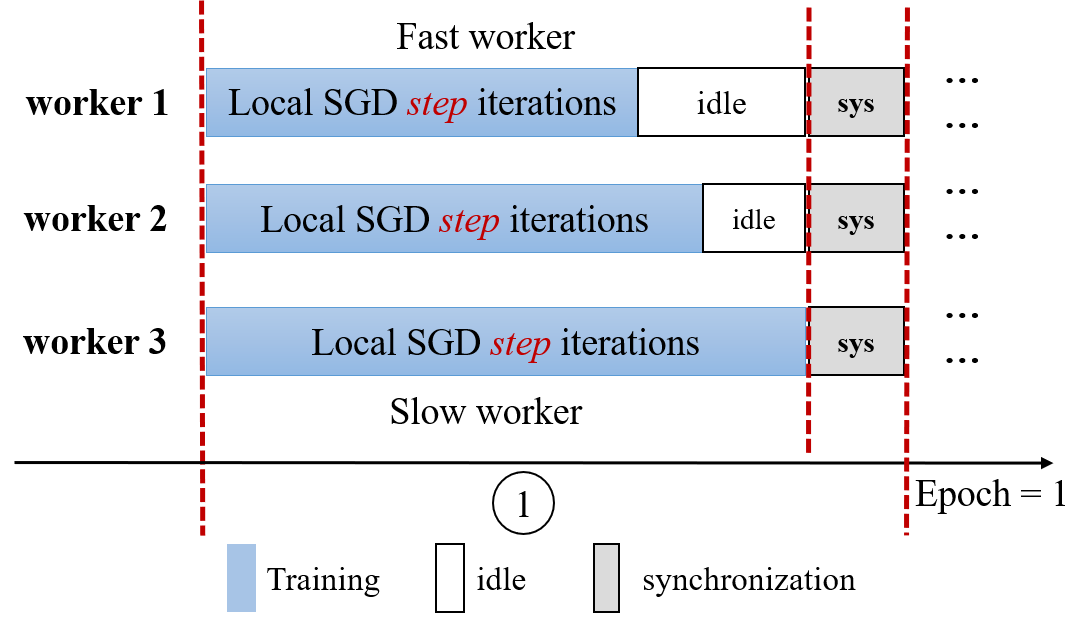}
           \caption{Fixed batch size}\label{fig:simulationa}
        \end{subfigure}%
        \hspace*{\fill}   % maximize separation between the subfigures
        \begin{subfigure}{0.45\textwidth}
            \includegraphics[width=\linewidth]{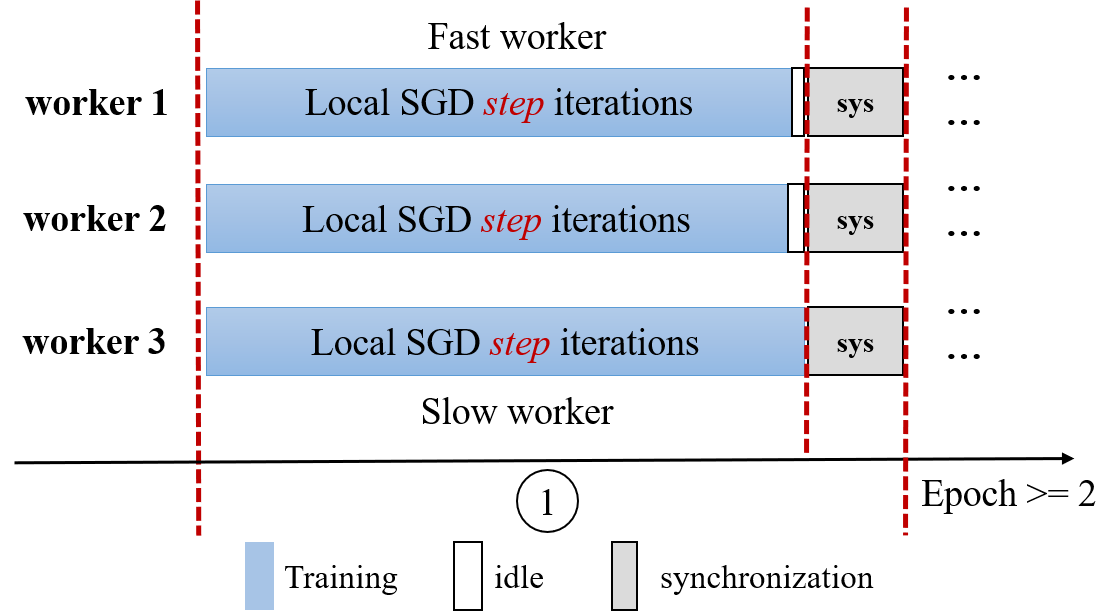}
            \caption{Dynamic batch size}\label{fig:simulationb}
        \end{subfigure}%
        \caption{The training simulation in an illustrative cluster of the synchronous methods with fixed batch size and dynamic batch size, where $step$ represents the synchronization interval, $step \geq 1$.}
        \label{fig:training simulation}
\end{figure}

    For example, an illustrative cluster consists of three workers as shown in Figure~(\ref{fig:simulationa}), which is utilized to train a DNN. The worker $1$ possesses the strongest performance, while worker $3$ has the weakest performance. Particularly, the workers $1$ and $2$ are idle when they are waiting for the worker $3$ after an iteration in a training epoch, which decreases the utilization of the cluster. The whole consumed time of training a DNN is formulated:~$T_a = t_{gpu} + t_w + t_s$, where $t_{gpu}$ indicates the consumed time of GPU, $t_w$ represents the waiting time because of the idle, $t_s$ indicates the synchronization time because of the communication among the cluster, respectively. As the training epoch increases, $t_w$ will be accumulated, which implies the waste of computational resource aggregates. 
    
    To alleviate this problem, the model averaging methods ~\cite{PoveyParallel,F-SGD,adaptive-SGD} were proposed to reduce the overhead of the synchronization by averaging the parameters of the DNN periodically (e.g., typically every one or two minutes or synchronization interval $step > 1$) instead of each iteration. Although these algorithms work well in practice, the best synchronous interval is hard to capture and the convergence time of training the DNN is still affected by synchronization period. Besides, the asynchronous algorithms \cite{Keuper2015Asynchronous,Damaskinos2018AsynchronousBM,APSGD} were also proposed to decrease the training time and improve the fault tolerance without a limit. However, it introduced an additional stale-gradient problem, which might result in the non-convergence of the DNN~\cite{Giladi2020AtSE}. In addition, some other state-of-the-art methods, concerning about decreasing the transmission cost of the synchronization are proposed, such as the gradient sparsification method~\cite{ShiA}, the quantification method~\cite{HubaraQuantized,WenXu17Terngrad}, and the compression method~\cite{LinDeep,2018arXiv180508768S}. However, these algorithms are based on the synchronization mechanism, which inherited the waste of resources caused by the different computational capabilities of the workers.
    
    We observe that the reason for the waiting among workers with different computational performance is the load imbalance of the distributed cluster. A novel Dynamic Batch Size~(DBS) strategy for distributed DNN training is proposed to guarantee the load balance of the cluster during the whole training of a DNN and thereby eliminating the waiting time. To the best of our knowledge, this is the first work focusing on the dynamic partitioning of the dataset rather than revising the synchronous or asynchronous methods. To achieve this, firstly, the performance evaluation of each worker is computed based on the consumed time and the sub-dataset in the previous epoch. Secondly, the batch size of each work and the partition of the dataset are dynamically adjusted according to the evaluated performance. Note that the estimated batch size and dataset partition will be utilized in the next training epoch. Finally, the dynamic adjustment of the batch size and dataset partition repeats until the distributed training of the DNN is finished. The goal of this paper is to discuss the potentiality and effectiveness of the proposed DBS strategy for distributed DNN training. The specific contributions of the proposed DBS algorithm are shown below:
    \begin{itemize}
        \item A novel DBS strategy for distributed DNN training is proposed to improve the utilization of the cluster on the basis of the worker performance, which is beneficial to boost the distributed DNN training.
        \item To the best of our knowledge, this is the first work that focuses on the robustness verification of the distributed DNN training with disturbance, which could be employed in all synchronous methods in principle. In addition, the essential theoretical analysis of the DBS algorithm is also provided.
        \item Extensive experiments are conducted and the experimental results indicate that the proposed DBS strategy outperforms the synchronous methods with fixed batch size in terms of the effectiveness and robustness. The source codes are released for the reproducibility~\footnote{\url{https://github.com/Soptq/Dynamic_Batch-Size_DistributedDNN}}. 
    \end{itemize}
    
    The rest of this paper is organized as follows. Related work is reviewed in Section~\ref{section:two}. Section~\ref{section:three} documents the details of the proposed DBS algorithm. The experiment design and the result analysis are shown in Section~\ref{section:four}, and the conclusions are drawn in Section~\ref{section:five}.

\section{Related work}
    \label{section:two}
    
    \subsection{Synchronous SGD with data parallelism}
    
    Synchronous SGD (S-SGD) with mini-batch is the most popular solution of the synchronous algorithms for distributed DNN training because of its satisfying reliability, stability, and convergence. Specifically, each worker hosts a copy of the large-scale DNN and an even dataset partition, and then the worker takes a mini-bath SGD at each iteration and replaces its solution by the average of all solutions of workers. Particularly, the consumed time of the distributed DNN training can be formulated as: $T_a = t_{gpu}+t_w+t_s$. The overview framework of the S-SGD is shown in Algorithm~\ref{alg:S-SGD}.
    \begin{algorithm}[h] 
        \caption{S-SGD with data parallelism} 
        \label{alg:S-SGD} 
        \begin{algorithmic}[1] 
            \State Initialization: $w_i^0$, learning rate: $\gamma \ge 0$, synchronization interval (integer): $step = 1$, max training: $epoch$, fixed mini-batch size: $b$;
            \label{ssgd:step1}
            \State Divide the dataset evenly:$D_i = \frac{D}{n}$, and calculate the iterations:~$T$;
            \label{ssgd:step2}
            \For {$e=0$ to $epoch$}
                \For {$t=0$ to $T$}
                \State Train the DNN with a mini-batch on each worker: $ compute(\nabla w_i^t)$;
                \label{ssgd:step3}
                \State Synchronize gradients:~$aggregate(\nabla w_i^t)$;
                \label{ssgd:step4}
                \State Update parameters: $w_i^{t+1}$ = $w_i^{t}- \frac{\gamma}{n}  \sum_{i=1}^n {\nabla w_i^t}$;
                \label{ssgd:step5}
                \State Redistribute $w_i^{t+1}$;
                \label{ssgd:step6}
                \EndFor
            \EndFor
        \end{algorithmic} 
    \end{algorithm}
    
Specifically, the dataset is evenly divided, and each computational worker is configured with the same hyper-parameters for the DNN training (e.g., fixed batch size, $\gamma$, optimizer) at Step \ref{ssgd:step1}. Then each worker computes the local gradient independently and all gradients are accumulated by well-designed methods. Afterward, the new parameters are calculated and redistributed. Particularly, two architectures, i.e., centralized architecture~\cite{li2014scaling,li2014communication} and decentralized architecture~\cite{DASGD,Li2018PipeSGDAD}, are designed to fulfil the distributed DNN training with data parallelism. However, the consumed time of the waiting and synchronization during the DNN training (i.e., the $t_w$ and $t_s$) are inevitable due to the different performance and synchronization operations at each iteration.

\textbf{Model Averaging:}~With a motivation to reduce the number of synchronization among inter-workers, the model averaging method~\cite{PoveyParallel} has been proposed and widely used in practical distributed training of DNNs, where the models on different workers are averaged after several iterations (i.e., $step \ge 1$). Particularly, If $step = 1$, it is the fully S-SGD, while it is referring to one-shot averaging~\cite{PSGD} if the averaging operation occurs only at the end of the training. Consequently, the model averaging methods decrease the synchronization cost by reducing the number of synchronization rounds in the training, thereby decreasing the $t_w$ and $t_s$. Extensive experiments~\cite{PRSGD,EMA} have demonstrated that the model averaging can reduce synchronization overhead of the training time as long as the period is suitable. In addition, some theoretical studies~\cite{PRSGD,F-SGD,2015arXiv150601900A,Parallel-SGD} have given the analysis of the reason for the model averaging achieving good convergence rate. However, the model averaging might cause the bias of the accuracy, since it suffers from a residual error with respect to fully synchronous SGD. Meanwhile, the understanding of how averaging period can affect the performance of parallel SGD is quite limited in the current literature.
    
\subsection{Asynchronous SGD}
To reduce the waiting time:~$t_{w}$, another natural idea is to relax the strict synchronization requirement. That is, by allowing the updates $\nabla w^t$ to be applied to calculate the global parameters as soon as they are computed (instead of waiting for synchronization of all workers). This method is named as Asynchronous SGD (A-SGD)~\cite{Keuper2015Asynchronous,HOGWILD} and it is widely utilized as an efficient distributed training strategy since it iterates faster without synchronization. However, although A-SGD offers a fast iteration speed, the gradients some workers used to update the model is stale. To be specific, assuming at iteration $t$, worker $i$ uses the global weight $w^t$ to starts a new iteration, then calculates the gradients ${\nabla} w^{t}$ and finally updates the model. By the time when worker $i$ finishes the iteration $t$, the global weight $w^t$ has already been updated $\tau$ times by other faster workers, becoming $w^{t+\tau}$. Hence, the update formula of the A-SGD is defined as Equation~\ref{equation1}. 
    \begin{equation}
    w^{t+\tau+1} = w^{t+\tau} - \gamma \cdot {\nabla}w_i^t
    \label{equation1}
    \end{equation}
    
    It can be seen that the formula produces a delay of $\tau$ steps compared to the S-SGD, which eventually leads to a loss of accuracy.
    
    To conquer the data staleness challenge, the method of bounded stale updates \cite{ho2013more,cipar2013solving} is proposed to suggest that fast workers can utilize the stale gradients to update the parameters of the model if the staleness is bounded below a limitation. However, this method introduces a bound-limitation parameter as a new hyper-parameter which is capable of affecting the accuracy of the model in the deep learning procedure. On the other hand, some variants of A-SGD~\cite{ASGDwithDC,2015Staleness-aware} are proposed to calculate the current global average gradients based on the staleness and stale gradients. Yet, in their experiments, although the accuracy of their methods is much improved compared to A-SGD, they still have some accuracy loss compared to S-SGD. Thus, as far as we know, all proposed methods that aim to eliminate $t_{w}$ by utilizing the asynchronous operation will also reduce the accuracy of the model on the corresponding problem to some degrees.
    \section{The proposed DBS algorithm}
    \label{section:three}
    The proposed DBS algorithm focuses on the reconstruction of Step~\ref{ssgd:step2} of Algorithm~\ref{alg:S-SGD}, which is composed of three components: 1) \textbf{Performance evaluation of the computational worker}. 2) \textbf{Dynamic adjustment of the batch size.} 3) \textbf{Dynamic partition of the dataset}, which will be detailed in the following sections. Specifically, Figure~\ref{fig:training simulation}(b) shows the distributed DNN training simulation that the DBS algorithm hopes to achieve. Although the computational capability of the workers is uneven, the DBS algorithm can also keep the load balance of the cluster so that $t_{gpu}$ of the different workers is approximately the same compared to the Figure~\ref{fig:training simulation}(a), and the waiting time:~$t_w$ is eliminated.

    \subsection{Performance evaluation of the computational worker}
    The performance of different workers in a cluster is affected by many factors (e.g, hardware, software, or temperature), and the relationship between the factors and performance is hard to capture. Moreover, some of these factors are random and not possible to infer (e.g., the GPU resource of a worker is consumed by a new irrelevant task). Hence, it's difficult to predict the performance of the worker accurately in the distributed training environment by gathering all information (e.g, GPU model or parameter size) and feeding them into a magic algorithm. 
    
    In this paper, an assessment method is proposed to evaluate the computational performance of the worker based on the assumption that the performance can be considered stable in a short period of time. Particularly, the current performance of a worker is estimated in dividing the proportion of sub-dataset $d_i^j$ (i.e., $d_i^j = \|D_i^j\|/\|D\|, \sum_i^n d_i^j = 1$) by the training time $t_{i}^j$ of the previous training epoch. Especially, the performance evaluating formula is illustrated as $p_i^j = d_i^j/t_i^j$, where $i$ and $j$ indicates the $i$-th worker and $j$-th epoch, respectively. After each training epoch, $p_i^j$ needs to be recalculated and will be utilized to adjust the batch size and dataset partition for the training in the next epoch. If $j=0$, the partition of the dataset is evenly divided and the batch size is fixed, just like S-SGD.
    \subsection{Dynamic adjustment of the batch size}
    In order to reasonably assign the load balance during the distributed training, the batch size of each worker should be dynamically adjusted based on its current performance which is represented by $p_i^j$. Then the adjusted batch size will be applied to the next epoch training to ensure the workers with different performance can finish their own tasks as close as possible. 
    
    \begin{algorithm}[h] 
        \caption{Dynamic Batch Size Algorithm} 
        \label{alg:calculatingBatchsize} 
        \begin{algorithmic}[1] 
            \Require 
            The performance evaluation of the $worker_i$ : $p_i^j$;
            \Ensure 
            The range of $worker_i$'s dataset :$(L^{j+1}, K^{j+1})$;
            \State $p_{sum} = \sum_1^n p_i^j $;
            \label{DBS:step1}
            \For {$i=1$, $n$}
            \State $b_i^{j+1} = \frac{p_i^j}{p_{sum}} $;
            \label{DBS:step2}
            \State $B_i^{j+1} = b_i^{j+1} \times B$;
            \label{DBS:step3}
            \EndFor
            \State $[L^{j+1}, K^{j+1}] = DynamicDatasetAdjust([B_1^{j+1}, B_2^{j+1}, B_i^{j+1}, \cdots, B_n^j])$; \\
            \label{DBS:adjustDataset}
            \Return $[L^{j+1}, K^{j+1}]$;
        \end{algorithmic} 
    \end{algorithm}
    
 The detail of the DBS strategy is shown in Algorithm \ref{alg:calculatingBatchsize}. Each worker gathers performance evaluation of all workers in the cluster by using $AllReduce$ operation~\cite{zhao-Kylix} at Step~\ref{DBS:step1}, which is an efficient way to fulfill the communication of the cluster. Then the dynamic batch size ratio $b_i^j$ of worker $i$ is computed with the performance of the worker $i$ at Step \ref{DBS:step2}, as the workload allocated to the worker is proportional to the production of its performance and train time, where the train time of all workers are expected to be approximate. Meanwhile, throughout the whole DNN training procedure, the total batch size $B$ of the whole cluster is fixed, and therefore the dynamic batch size of each worker can be presented as $B_i^{j+1} = b_i^{j+1} \times B$. At last, the partition of the dataset is also adjusted by $DynamicDatasetAdjust$ function based on the new batch size $B^{j+1}$. The outputs of $DatasetAdjust$ is $L^j$ and $K^j$, indicating the range of the sub-dataset for each worker. 
    \subsection{Dynamic partition of the dataset}
    The dynamic partition of the dataset is approximate with the ratio of each worker's batch size. The batch size represents the sample size of an iteration and should be an integer, while $B_i^{j+1}$ is very likely to be a non-integer. Hence, $B_i^{j+1}$ needs to converse to the integer $B_i^{'j+1}$ and then used to calculate the partition of the dataset. Particularly, $B_i^{'j+1}$ should satisfy the conditions:
    \begin{equation}
    \sum_i^n B_i^{'j+1} \leq B \label{equ:5}
    \end{equation}
    \begin{equation}
    \min[\sum (B_i^{'j+1} - B_i^{j+1})^2] \label{equ:6}
    \end{equation}
    
     To achieve this, a method of rounding twice is proposed. Firstly, each initial batch size $B_i^{j+1}$ is rounded down to $B_i^{('j+1)}$ (i.e., $B_i^{('j+1)} = \lfloor{B_i^{j+1}}\rfloor$). The difference between the initial sum of adjusted batch size and the total batch size can be represented as: $k = B - \sum_i^n B_i^{'(j+1)}$, indicating there are at most $k$ values that can be rounded up. Secondly, in order to satisfy the equation \eqref{equ:6}, $B^{j+1}$ will be sorted in the decimal descending order (i.e., $argsort(B_{decimal}^{j+1})$, $B_{decimal}$ presents the decimals fraction of the $B^{j+1}$) , then top-$m$ values are picked to round up, where $m \leq k$. Specifically, the indexes of the top-$m$ Batch size is fetched (i.e., $id_1$, $id_2$,$……$, $id_m$) and $B_{id_m}^{('j+1)} = B_{id_m}^{('j+1)} +1$, which represents the rounding up of $B_{id_m}^{(j+1)}$. Furthermore, the value's decimal fraction of the $B_{id_m}^{(j+1)}$ should be greater or equal than $0.5$. Finally, $B^{'j+1}$ is normalized, and the ratio of $B_i^{'j+1}$ is viewed as the adjusted proportion of the sub-dataset approximately. The pseudo-code and the example of the $DynamicDatasetAdjust$ is detailed in supplementary.
    
    \subsection{Convergence analysis} 
    \label{Theoretical}
    
    Roughly speaking, the proposed DBS is a parallel synchronized mini-batch SGD, whose batch size is a variable. The variance of the batch size brings a new noise in the optimization process, which is negligible as demonstrated in this section. Specifically, the convergence analysis of DBS consists of two parts: general synchronized mini-batch SGD modeling, and analysis of DBS.
    
    Considering giving a general analysis of discussing the convergence of mini-batch S-SGD, The scenario of parallel mini-batch S-SGD can be modeled as the following optimization problem:
    \begin{equation} 
    x^{*}=\mathop{\arg\min}_{x \in \mathbb{R}^{d}} f(x)
    \label{ta:1}
    \end{equation} 
    where $f(x) =\frac{1}{n}\sum_{i=1}^{n} f_{i}(x)$, and each $f_{i}:\mathbb{R}^{d} \rightarrow \mathbb{R}$ is smooth and $f(x)$ is $\mu$-strongly-convex, which satisfy:
    \begin{equation}
    f(x^*) \ge f(x) + \langle \nabla f(x), x^{*}-x \rangle + \frac{\mu}{2}\|x^{*}-x\|^{2} 
    \label{ta:2}
    \end{equation}
    where $x^{*}$ is assumed the unique global minimizer.
    For each $f_{i}(x)$ obviously we have:
    \begin{equation}
    \mathbb{E} [ f_{i}(x) = f(x) ] \label{ta:3}
    \end{equation}
    \begin{equation}
    \mathbb{E} [ \nabla f_{i}(x) ] = \nabla f(x) \label{ta:4}
    \end{equation}
    By iteration $j$, using SGD, step-size $\gamma$ \textgreater~0:
    \begin{equation}
    x^{j+1} = x^{j} - \gamma \nabla f_{i}(x^{j})
    \label{ta:5}
    \end{equation}
    
    Particularly, the mini-batch size $b_{i}$  of the proposed DBS strategy turns into random variable from scalar by altering $b_{i}$ and rounding off, which cause the variability. $b_{i}$ could be described in terms of its expectation $\mathbb{E}(b_{i})$ and variance $\sigma_{b}^2$.
    
    May as well suppose $\mathbb{E} [ \| \nabla f_{i}(x) \| ^{2}] \le G^2$. Considering $\| \nabla f_{i}(x) \|$ related to the size of a anisotropic-circular field where $\nabla f_{i}(x)$ might land, $\sigma_{i}=\sigma(f_{i},b_{i})$ the deviation related to the field could be assume to be finite (considering $\sum b_{i} \in [B-n,B]$ and Lemma~\ref{lm1} in supplementary).
    
    In the DBS algorithm, gradient noise $\sigma_{i}^{*}$, composed of the noise from dynamic mini-batch ($\sigma_{i}$) and inherent one ($\hat{\sigma}_{i}$). In a tricky way, the $\sigma$ is assumed to be an upper bound of all the gradient noise $\sigma_{i}^{*}$ and we could achieve:
    \begin{equation}
    \mathbb{E} [ \| \nabla f_{i}(x) \| ^{2} ] \leq \sigma ^{2}
    \label{ta:6}
    \end{equation}
    where $\sigma \leq \sigma_{min}$, as $\sigma_{min}$ to be the S-SGD with fixed mini-batch size $\mathop{min}\{b_{i}^{j}\}$. To figure out the same deduction, \cite{gower2019sgd} use expected smoothness~\cite{gower2018stochastic} and a weak assumption to deal with it.
    \newtheorem{thm}{\bf Theorem}
    \begin{thm}\label{ta:t1}
        Assume $f$ is $\mu$-strongly-convex and each $f_{i}$ has a gradient noise upper bound $\sigma ^{2}$ . Choose $\gamma \in (0, \mu^{-1})$ then the SGD iterates given by (\ref{ta:5}) satisfy the Equation~\ref{ta:7} (the proof~\ref{prf:ta:t1} is provided in supplementary):
        \begin{equation}
        \mathbb{E} \| x^{k}-x^{*} \| ^{2} \le (1-\gamma\mu)^{j} \| x^{0} - x^{*} \| ^{2} + \frac{\gamma\sigma^{2}}{\mu}
        \label{ta:7}
        \end{equation}
    \end{thm}
    
    Further, multitudinous methods could be adopted to control the step-size  $\gamma$, such as reducing $\gamma$ when $j$ increases, making the upper bound lower and even zero. The dynamic mini-batch size produces another gradient noise. However, it could be eliminated in some simple way. 
    
    In summary, the convergence of DBS may be the same as S-SGD taking $\mathop{min} \{ b_{i}^{j} \}$ as fixed mini-batch size, and even better, because of smaller the size, larger the noise. In practice, dynamic size tends to be stable so long as the cluster is stable, and the performance of the DBS strategy is expected to be the same as S-SGD with fixed mini-batch size.
    \section{Experiments}
    \label{section:four}
    \textbf{Experimental Setup.} All experiments are conducted on a single Tesla V100 machine with 4 GPUs. Multiple processes are employed to simulate the illustrative clusters in the distributed environment, i.e., each process runs as a computational worker. S-SGD is chosen as the baseline since it owns zero gradient staleness and achieves the best model accuracy. More experimental results (e.g., comparison between DBS and model averaging methods) are provided in supplementary. Specifically, the communication mechanism of the cluster adopts the $AllReduce$ approach~\cite{zhao-Kylix}. The ResNet101~\cite{he2016deep} with forty million parameters is chosen as a case of the large-scale DNN to validate the effectiveness of the DBS method, and the CIFAR10~\cite{CIFAR-10} is chosen as the benchmark datasets.
    
\begin{figure}[!h]
        \centering
        \begin{subfigure}{0.455\linewidth}
            \centering \includegraphics[width=\linewidth]{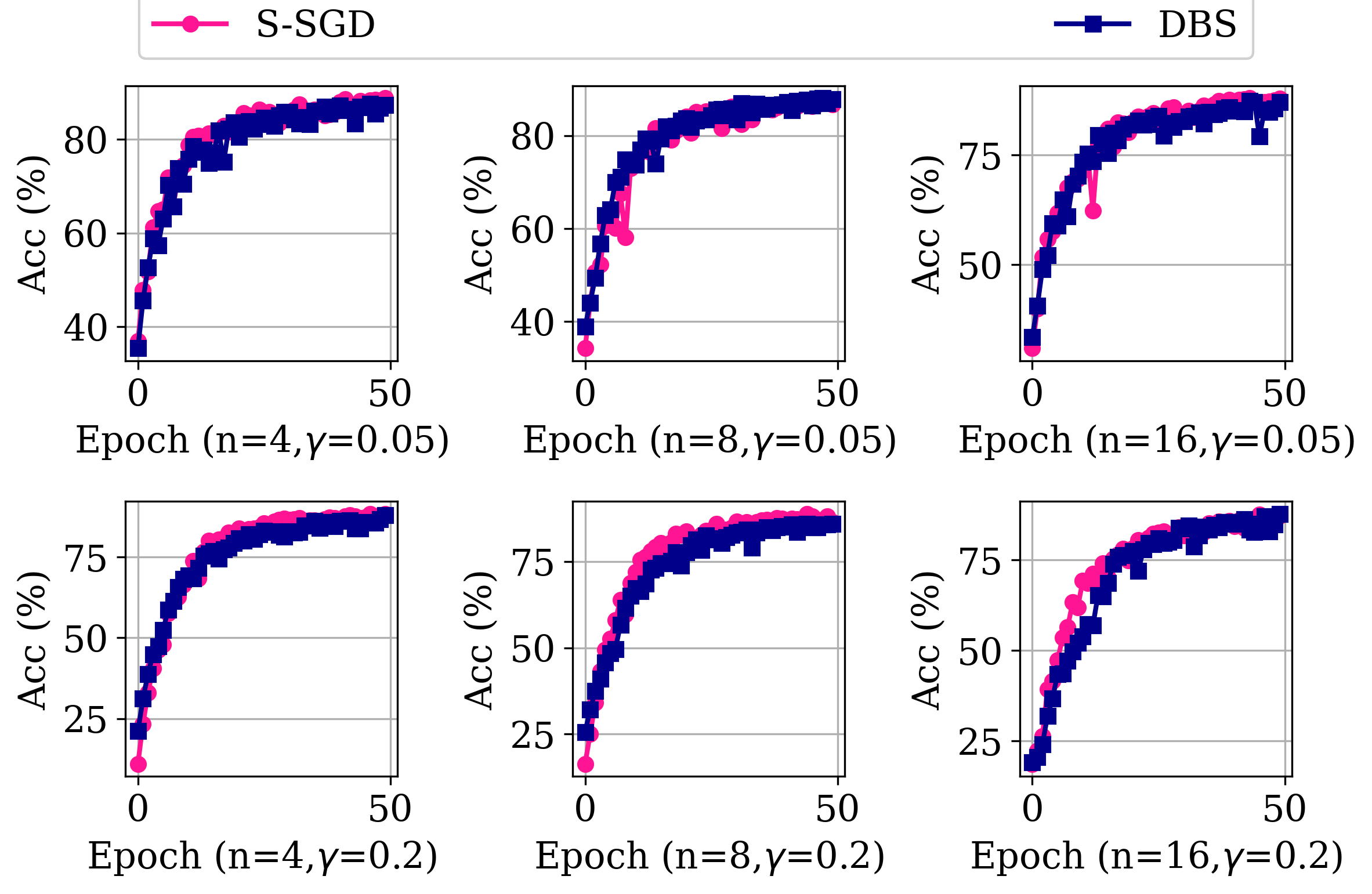}
        \end{subfigure}
        \begin{subfigure}{0.513\linewidth}
            \centering
            \includegraphics[width=\linewidth]{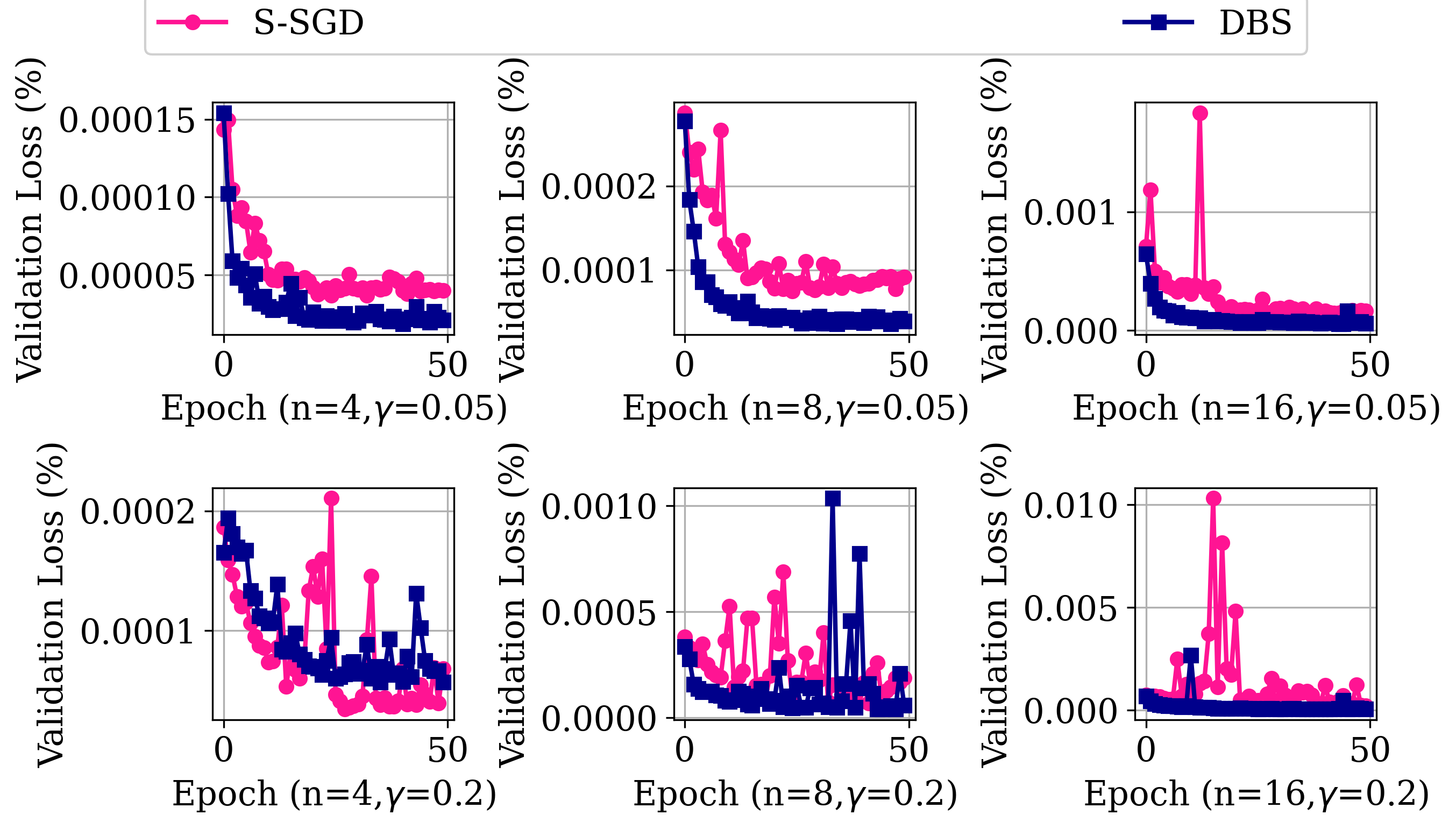}
        \end{subfigure}
        \caption{The accuracy and loss comparison between S-SGD and DBS with different cluster scales and initial learning rate.}
        \label{fig:ac}
    \end{figure}

   \textbf{Convergence comparison.} The convergence of the proposed DBS algorithm has been proved to be nearly consistent with the classical dense S-SGD, thus we will validate the theory with experimental results. In order to obtain the results in an acceptable training time, we set $epoch$-$size$ = 50, initial learning rate $\gamma = 0.05, 0.2$, and $batch$-$size$ = 512. To improve the convergence, as mentioned in Section. \ref{Theoretical}, $momentum$ = 0.5 is configured in both DBS and S-SGD implementation and Loss function is cross-entropy. Note that all subsequent experiments share the same experimental setting. The comparison of the training accuracy and validation loss between DBS and S-SGD will be conducted on different cluster scales, and the results are presented in Figure~\ref{fig:ac}. 
    
    According to Theorem \ref{ta:t1}, the impact of the gradient noise $ \sigma_{i} $ from dynamic mini-batch decreases as $\gamma$ obtains correction. Particularly, Figure~\ref{fig:ac} shows the results on accuracy with respect to training epochs. we see that they converge at basically the same accuracy on different cluster scales and learning rate, and both the fitting curves almost got the same slope at each the same epoch, which matches the theoretical analysis of DBS convergence.

\begin{figure}[!h]
        \centering
        \begin{subfigure}{0.31\linewidth}
            \centering
            \includegraphics[width=\linewidth]{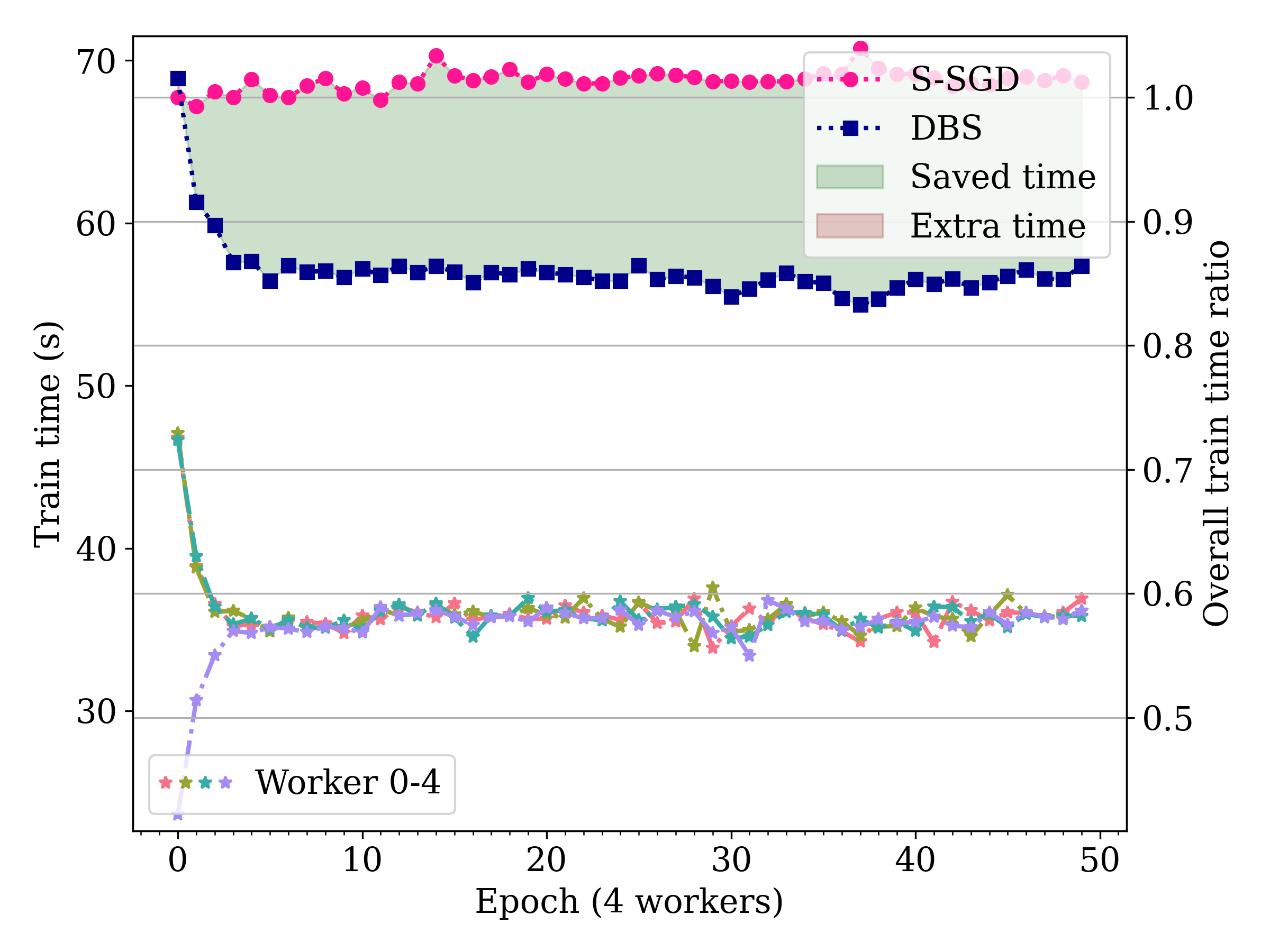}
        \end{subfigure}
        \begin{subfigure}{0.31\linewidth}
            \centering
            \includegraphics[width=\linewidth]{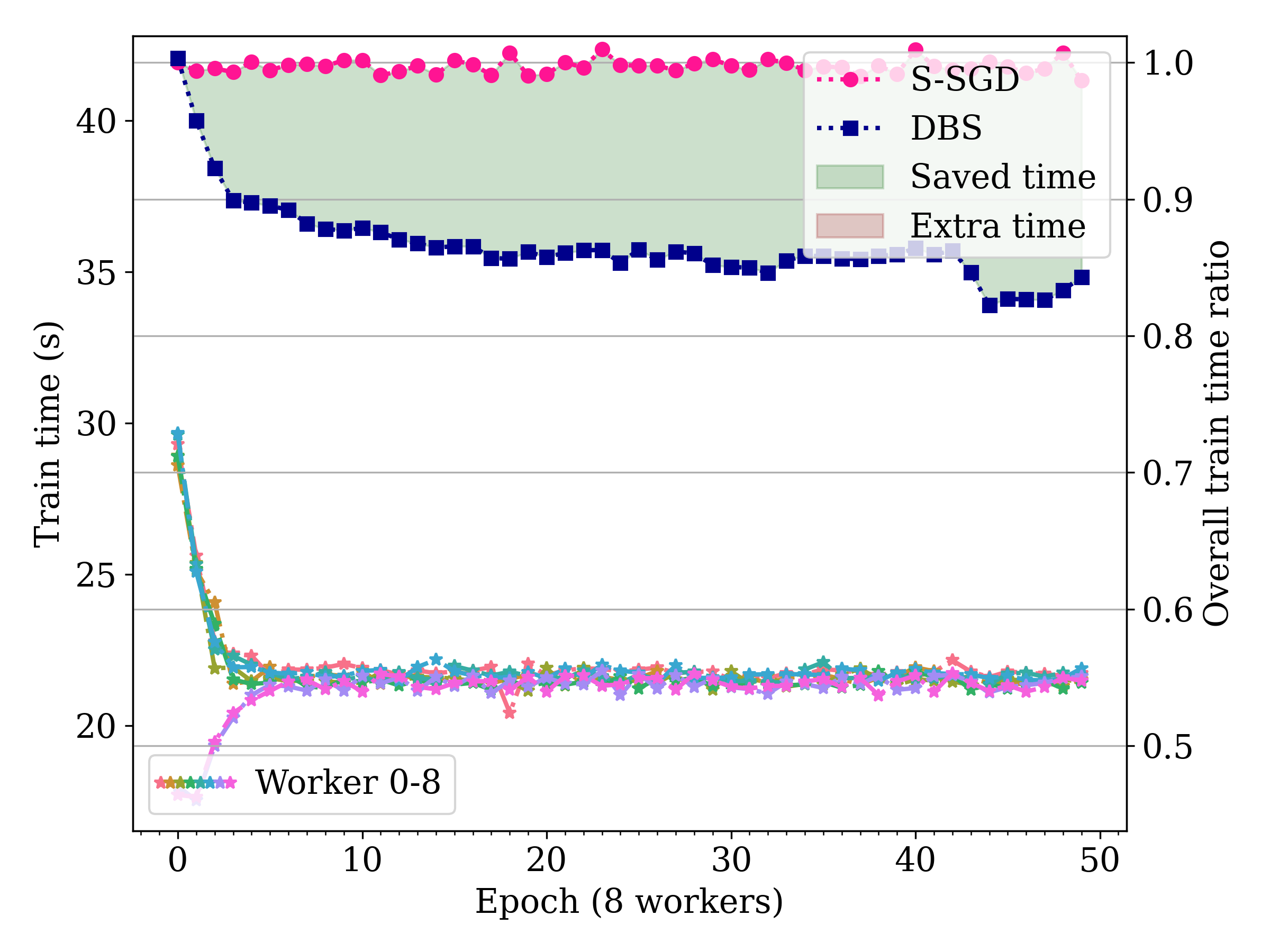}
        \end{subfigure}
        \begin{subfigure}{0.31\linewidth}
            \centering
            \includegraphics[width=\linewidth]{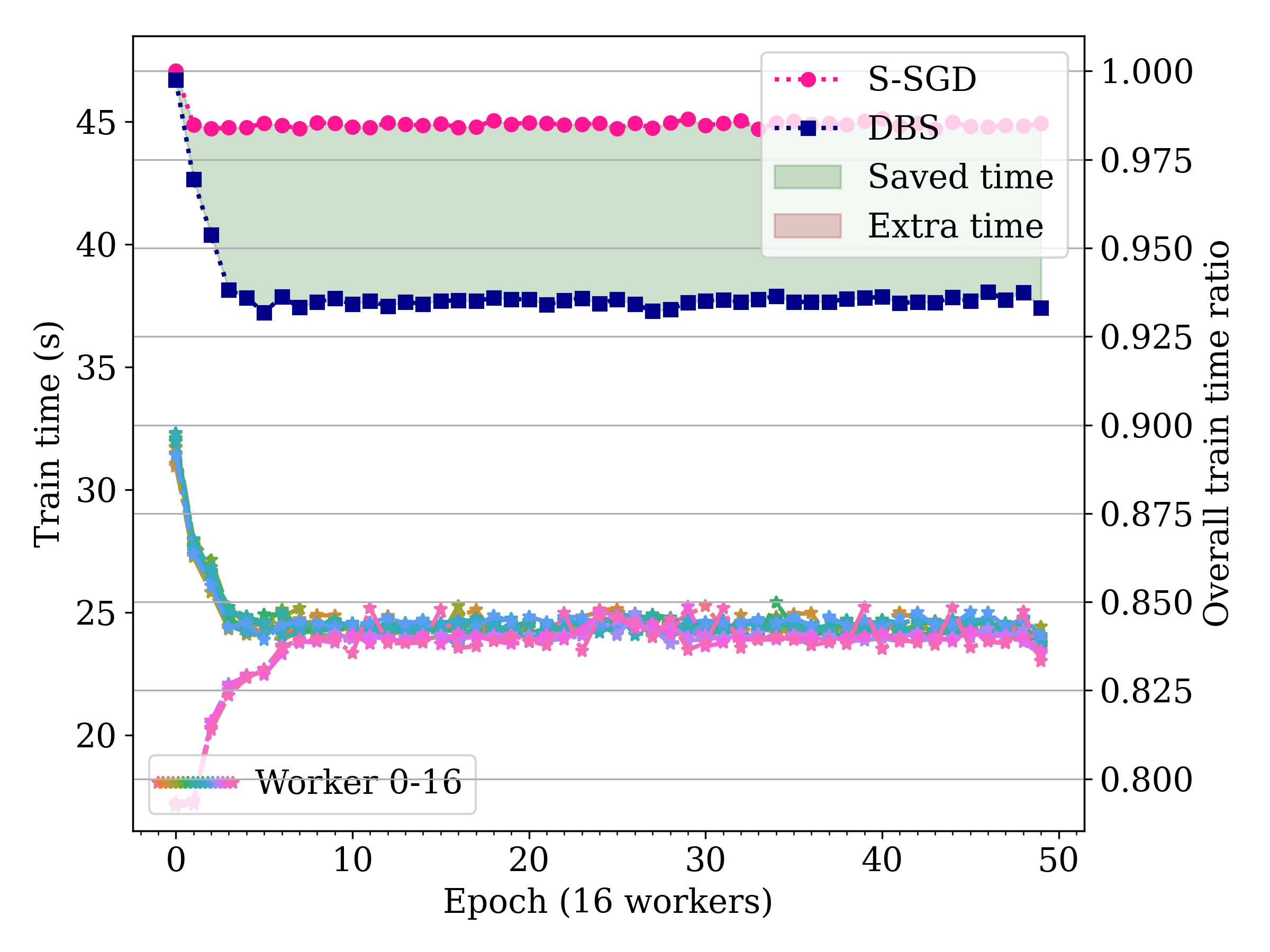}
        \end{subfigure}
        \caption{The top half of the figures represents comparison of the whole consumed time (i.e., $T_a$) between the BBS and S-SGD with respect to the epoch, while the bottom half represents the comparison of the consumed GPU time~(i.e, $t_{gpu}$)   }
        \label{fig:consumed-time}
    \end{figure}

 \textbf{Effectiveness and expansibility of the DBS.} We will investigate the effectiveness of DBS by training the ResNet101 on CIFAR10 with different scales. The whole time (i.e., $T_a = t_{gpu}+t_w+t_s$) and GPU time (i.e, $t_{gpu}$) of each epoch consumed by DBS and the typical S-SGD with different cluster scales are collected and presented in Figure~\ref{fig:consumed-time}, respectively.
    
 As shown in Figure~\ref{fig:consumed-time}, the consumed GPU time of the cluster is various at the first few epochs. As the training forwards, the $t_{gpu}$ becomes approximately the same. The main reason is that the computational performance of the workers is different and the partition of the dataset is even, thus the gap in GPU time consumed by different workers is wide at the beginning, which brings the synchronous overhead and results in the waste of the computational resource. After the dynamic adjustment of the batch size and dataset by the DBS strategy, the gap of $t_{gpu}$ is eliminated or narrowed and the waste of the computational resources is decreased, which is of great benefit to the utilization of the cluster. Therefore, the whole training time of one epoch with DBS strategy is less than S-SGD's as shown in Figure~\ref{fig:consumed-time} and the surface between two lines (i.e., $T_a$ of the DBS and S-SGD, respectively) donates the saved time. When the scale $n$ is $4$, the whole consumed time of each epoch was saved approximately 12\% compared to the baseline, but the advantage of the DBS decreases with the increase of the scale of the cluster. For example, when the scale $n$ is $8$, the saved time decreases to 10\%, and it decreases more at $n = 16$. The one reason is the cluster maintenance cost (i.e., synchronization and communication cost) increases with the expansion of the cluster, which reduces the proportion of GPU time in the total training time. The other is the cluster expansion leads to fine-grained task division, which narrows the gap of the training time. In summary, the DBS strategy can effectively maintain the load balance of the cluster, speed up the distributed training by improving the utilization of the cluster, and have good expansibility simultaneously.

\begin{figure}
        \begin{subfigure}{0.45\textwidth}
            \includegraphics[width=\linewidth]{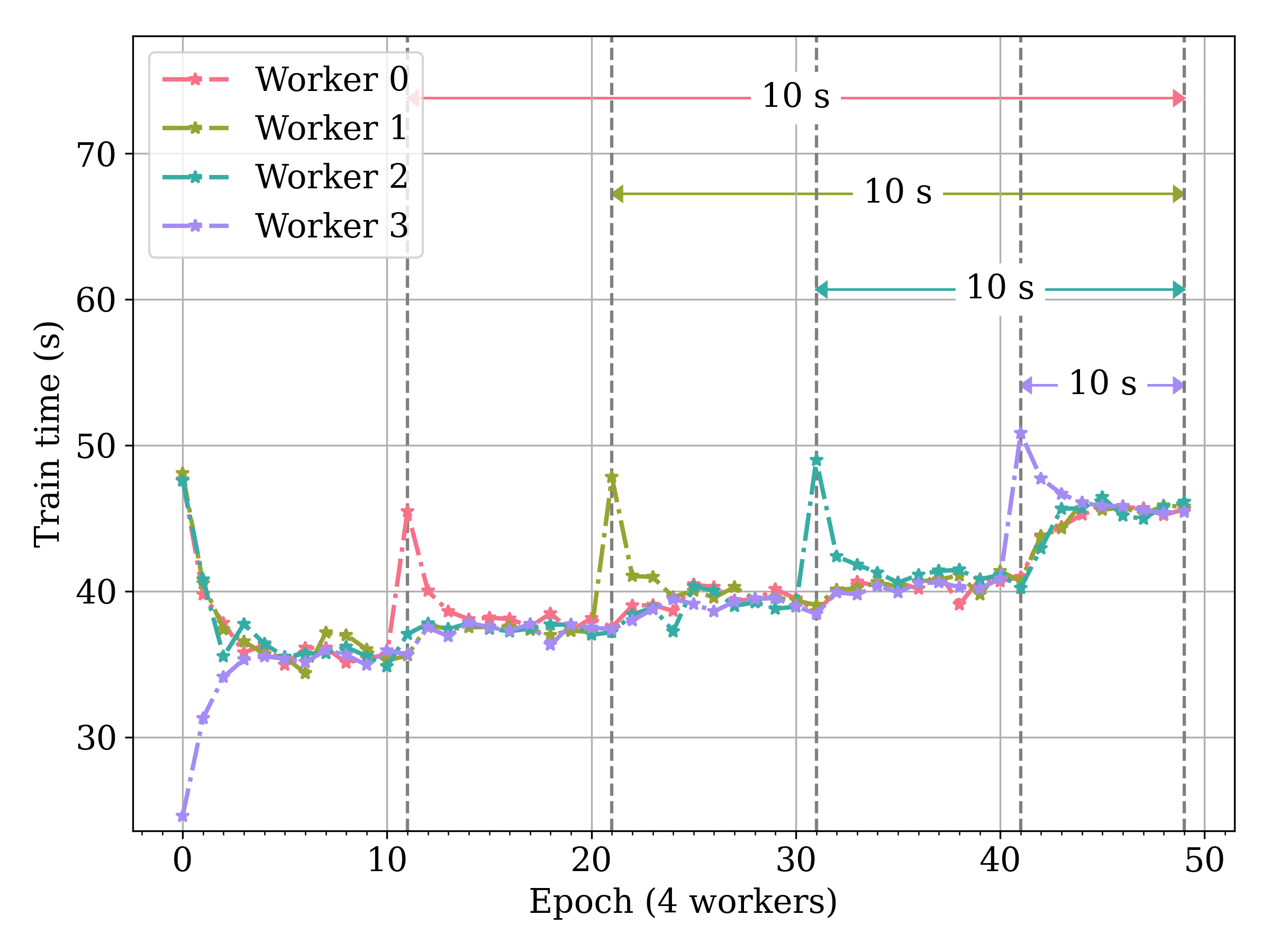}
            \caption{The dynamic GPU time of different workers with disturbance in DBS strategy.} \label{fig:robustnessa}
        \end{subfigure}%
        \hspace*{\fill}   % maximize separation between the subfigures
        \begin{subfigure}{0.45\textwidth}
            \includegraphics[width=\linewidth]{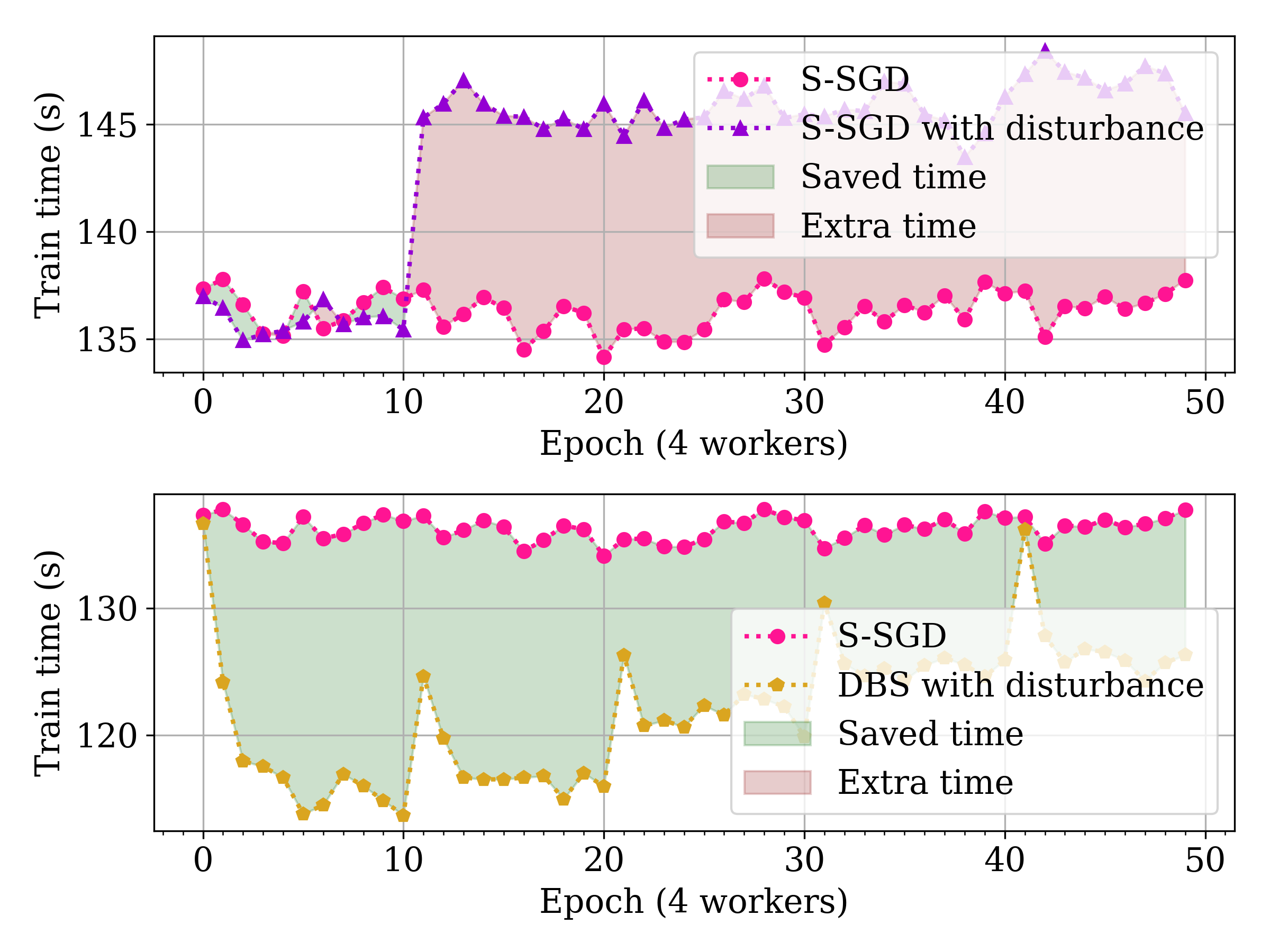}
            \caption{The whole consumed time comparison of the S-SGD and DBS with disturbance} \label{fig:robustnessb}
        \end{subfigure}%
        \caption{The robustness comparison of the S-SGD and DBS}
    \end{figure}

 \textbf{Robustness of the DBS.} We will observe the robustness of the proposed DBS strategy by running other irrelevant tasks periodically during the distributed training on an illustrative cluster with 4 computational workers. The consumed time of between DBS and S-SGD with respect to the training epochs are summarized in Figure~\ref{fig:robustnessb}.
    
    % \begin{figure}[h]
    %     \centering
    %     \includegraphics[width=0.92\linewidth]{PIC/disturbance.png}
    %     \caption{(a)The dynamic GPU time of different workers with disturbance in DBS strategy. (b) The robustness comparison of the S-SGD and DBS with disturbance}
    %     \label{fig:robustness}
    % \end{figure}
    
    Particularly, the occurrence of irrelevant task consumes the limited GPU resource, which leads to an increase in the consumed GPU time of one training epoch. For example, the consumed GPU time of workers is approximately the same during epochs 2-10 in Figure~\ref{fig:robustnessa}, which donates that the load balance is kept by the DBS. Suddenly, the training time of the worker 0 is more than others because of the disturbance at epoch 10 which increases the GPU time 10s until the distributed DNN training is over. That means the load balance is broken. Under the circumstance, the DBS strategy quickly adjusts the batch size and partition of the dataset to keep the balance, and the training time tends to be the same in the following training (e.g., the epochs 11-20). In practice, the disturbance occurs several times during the whole training process (e.g., the $t_{gpu}$ of the workers $1$ and $2$ suddenly increases at 21 and 31, respectively), and the DBS strategy works well. With the disturbance of the irrelevant task, the whole consumed time $T_a$ of the S-SGD increases rapidly because of the synchronization, while the DBS strategy has good robustness to overcome the disturbance and saves lots of time as shown in Figure~\ref{fig:robustnessb}. All experiments demonstrate that the DBS has good load balancing ability compared to the classical S-SGD, which is beneficial to improve cluster utilization and reduce training time.
    
    \section{Conclusions}
    \label{section:five}
    This paper proposed a novel Dynamic Batch Size (DBS) strategy for distributed DNN training, which makes the training time of the cluster in each iteration approximately the same by keeping the load balance of the cluster. The elimination of the waiting time among workers improves the utilization of the cluster and thereby boosts the distributed DNN training. The theoretical analysis of DBS is also provided. Extensive experimental results demonstrate the utilization and robustness of the DBS as it outperforms the classical synchronous methods with fixed batch size.

\section*{Broader Impact}
To the best of our knowledge, the proposed DBS strategy is the first work focusing on the robustness of the distributed DNN training with disturbance, which can be utilized to boost all synchronous methods such as S-SGD, the model averaging methods. In other words, the researchers of the distributed DNN training no longer only focus on the improvement of the algorithm, but the acceleration of the distributed training from the perspective of the dataset partitioning and load balancing. Particularly, the experimental results have shown that the convergence of DBS is approximately the  same with the S-SGD, while the DBS can effectively improve the utilization of the distributed cluster compared to the synchronous methods with good robustness.
\begin{ack}
	This work is supported in part by the State Key Program of the National Science Foundation of China under Grant 61836006.
\end{ack}
\bibliographystyle{plainnat}
\small\bibliography{neurips_2020.bib}

\begin{thebibliography}{37}
\providecommand{\natexlab}[1]{#1}
\providecommand{\url}[1]{\texttt{#1}}
\expandafter\ifx\csname urlstyle\endcsname\relax
  \providecommand{\doi}[1]{doi: #1}\else
  \providecommand{\doi}{doi: \begingroup \urlstyle{rm}\Url}\fi

\bibitem[{Arjevani} and {Shamir}(2015)]{2015arXiv150601900A}
Yossi {Arjevani} and Ohad {Shamir}.
\newblock {Communication Complexity of Distributed Convex Learning and
  Optimization}.
\newblock \emph{arXiv e-prints}, art. arXiv:1506.01900, June 2015.

\bibitem[Cheatham et~al.(1995)Cheatham, Fahmy, Stefanescu, and Valiant]{bsp}
Thomas Cheatham, Amr Fahmy, D.~C. Stefanescu, and L.~G. Valiant.
\newblock computing-a paradigm for transportablesoftware.
\newblock In \emph{System Sciences, 1995. Vol. II. Proceedings of the
  Twenty-Eighth Hawaii International Conference on}, 1995.

\bibitem[Cipar et~al.(2013)Cipar, Ho, Kim, Lee, Ganger, Gibson, Keeton, and
  Xing]{cipar2013solving}
James Cipar, Qirong Ho, Jin~Kyu Kim, Seunghak Lee, Gregory~R Ganger, Garth
  Gibson, Kimberly Keeton, and Eric Xing.
\newblock Solving the straggler problem with bounded staleness.
\newblock In \emph{Presented as part of the 14th Workshop on Hot Topics in
  Operating Systems}, 2013.

\bibitem[Damaskinos et~al.(2018)Damaskinos, Mhamdi, Guerraoui, Patra, and
  Taziki]{Damaskinos2018AsynchronousBM}
Georgios Damaskinos, El~Mahdi~El Mhamdi, Rachid Guerraoui, Rhicheek Patra, and
  Mahsa Taziki.
\newblock Asynchronous byzantine machine learning (the case of sgd).
\newblock In \emph{ICML}, 2018.

\bibitem[Dean et~al.(2012)Dean, Corrado, Monga, Kai, and Ng]{Dean2012Large}
Jeffrey Dean, Greg~S Corrado, Rajat Monga, Chen Kai, and Andrew~Y Ng.
\newblock Large scale distributed deep networks.
\newblock \emph{Advances in Neural Information Processing Systems}, 2012.

\bibitem[Giladi et~al.(2020)Giladi, Nacson, Hoffer, and Soudry]{Giladi2020AtSE}
Niv Giladi, Mor~Shpigel Nacson, Elad Hoffer, and Daniel Soudry.
\newblock At stability's edge: How to adjust hyperparameters to preserve minima
  selection in asynchronous training of neural networks?
\newblock \emph{ArXiv}, abs/1909.12340, 2020.

\bibitem[Gower et~al.(2018)Gower, Richtárik, and Bach]{gower2018stochastic}
Robert~M. Gower, Peter Richtárik, and Francis Bach.
\newblock Stochastic quasi-gradient methods: Variance reduction via jacobian
  sketching, 2018.

\bibitem[Gower et~al.(2019)Gower, Loizou, Qian, Sailanbayev, Shulgin, and
  Richtarik]{gower2019sgd}
Robert~Mansel Gower, Nicolas Loizou, Xun Qian, Alibek Sailanbayev, Egor
  Shulgin, and Peter Richtarik.
\newblock Sgd: General analysis and improved rates, 2019.

\bibitem[Haddadpour et~al.(2019)Haddadpour, Kamani, Mahdavi, and
  Cadambe]{adaptive-SGD}
Farzin Haddadpour, Mohammad~Mahdi Kamani, Mehrdad Mahdavi, and Viveck Cadambe.
\newblock Local sgd with periodic averaging: Tighter analysis and adaptive
  synchronization.
\newblock In \emph{Advances in Neural Information Processing Systems}, pages
  11080--11092, 2019.

\bibitem[He et~al.(2016)He, Zhang, Ren, and Jian]{he2016deep}
Kaiming He, Xiangyu Zhang, Shaoqing Ren, and Sun Jian.
\newblock Deep residual learning for image recognition.
\newblock In \emph{2016 IEEE Conference on Computer Vision and Pattern
  Recognition (CVPR)}, 2016.

\bibitem[Ho et~al.(2013)Ho, Cipar, Cui, Lee, Kim, Gibbons, Gibson, Ganger, and
  Xing]{ho2013more}
Qirong Ho, James Cipar, Henggang Cui, Seunghak Lee, Jin~Kyu Kim, Phillip~B
  Gibbons, Garth~A Gibson, Greg Ganger, and Eric~P Xing.
\newblock More effective distributed ml via a stale synchronous parallel
  parameter server.
\newblock In \emph{Advances in neural information processing systems}, pages
  1223--1231, 2013.

\bibitem[{Hubara} et~al.(2016){Hubara}, {Courbariaux}, {Soudry}, {El-Yaniv},
  and {Bengio}]{HubaraQuantized}
Itay {Hubara}, Matthieu {Courbariaux}, Daniel {Soudry}, Ran {El-Yaniv}, and
  Yoshua {Bengio}.
\newblock {Quantized Neural Networks: Training Neural Networks with Low
  Precision Weights and Activations}.
\newblock \emph{arXiv e-prints}, art. arXiv:1609.07061, Sep 2016.

\bibitem[Keuper and Pfreundt(2015)]{Keuper2015Asynchronous}
Janis Keuper and Franz-Josef Pfreundt.
\newblock Asynchronous parallel stochastic gradient descent - a numeric core
  for scalable distributed machine learning algorithms.
\newblock \emph{Computer Science}, page~1, 2015.

\bibitem[Krizhevsky and Hinton(2009)]{CIFAR-10}
A.~Krizhevsky and G.~Hinton.
\newblock Learning multiple layers of features from tiny images.
\newblock \emph{Computer Science Department, University of Toronto, Tech. Rep},
  1, 01 2009.

\bibitem[Li et~al.(2014{\natexlab{a}})Li, Andersen, Park, Smola, Ahmed,
  Josifovski, Long, Shekita, and Su]{li2014scaling}
Mu~Li, David~G Andersen, Jun~Woo Park, Alexander~J Smola, Amr Ahmed, Vanja
  Josifovski, James Long, Eugene~J Shekita, and Bor-Yiing Su.
\newblock Scaling distributed machine learning with the parameter server.
\newblock In \emph{11th $\{$USENIX$\}$ Symposium on Operating Systems Design
  and Implementation ($\{$OSDI$\}$ 14)}, pages 583--598, 2014{\natexlab{a}}.

\bibitem[Li et~al.(2014{\natexlab{b}})Li, Andersen, Smola, and
  Yu]{li2014communication}
Mu~Li, David~G Andersen, Alexander~J Smola, and Kai Yu.
\newblock Communication efficient distributed machine learning with the
  parameter server.
\newblock In \emph{Advances in Neural Information Processing Systems}, pages
  19--27, 2014{\natexlab{b}}.

\bibitem[Li et~al.(2018)Li, Yu, Li, Avestimehr, Kim, and
  Schwing]{Li2018PipeSGDAD}
Youjie Li, Mingchao Yu, Songze Li, Amir~Salman Avestimehr, Nam~Sung Kim, and
  Alexander~G. Schwing.
\newblock Pipe-sgd: A decentralized pipelined sgd framework for distributed
  deep net training.
\newblock In \emph{NeurIPS}, 2018.

\bibitem[Lian et~al.(2015)Lian, Huang, Li, and Liu]{APSGD}
Xiangru Lian, Yijun Huang, Yuncheng Li, and Ji~Liu.
\newblock Asynchronous parallel stochastic gradient for nonconvex optimization.
\newblock In C.~Cortes, N.~D. Lawrence, D.~D. Lee, M.~Sugiyama, and R.~Garnett,
  editors, \emph{Advances in Neural Information Processing Systems 28}, pages
  2737--2745. Curran Associates, Inc., 2015.
\newblock URL
  \url{http://papers.nips.cc/paper/5751-asynchronous-parallel-stochastic-gradient-for-nonconvex-optimization.pdf}.

\bibitem[{Lian} et~al.(2017){Lian}, {Zhang}, {Zhang}, and {Liu}]{DASGD}
Xiangru {Lian}, Wei {Zhang}, Ce~{Zhang}, and Ji~{Liu}.
\newblock {Asynchronous Decentralized Parallel Stochastic Gradient Descent}.
\newblock \emph{arXiv e-prints}, art. arXiv:1710.06952, October 2017.

\bibitem[Lin et~al.(2018)Lin, Han, Mao, Wang, and Dally]{LinDeep}
Yujun Lin, Song Han, Huizi Mao, Yu~Wang, and Bill Dally.
\newblock Deep gradient compression: Reducing the communication bandwidth for
  distributed training.
\newblock In \emph{International Conference on Learning Representations}, 2018.
\newblock URL \url{https://openreview.net/forum?id=SkhQHMW0W}.

\bibitem[{Niu} et~al.(2011){Niu}, {Recht}, {Re}, and {Wright}]{HOGWILD}
Feng {Niu}, Benjamin {Recht}, Christopher {Re}, and Stephen~J. {Wright}.
\newblock {HOGWILD!: A Lock-Free Approach to Parallelizing Stochastic Gradient
  Descent}.
\newblock \emph{arXiv e-prints}, art. arXiv:1106.5730, June 2011.

\bibitem[{Ouyang} et~al.(2020){Ouyang}, {Dong}, {Xu}, and {Xiao}]{COS}
Shuo {Ouyang}, Dezun {Dong}, Yemao {Xu}, and Liquan {Xiao}.
\newblock {Communication Optimization Strategies for Distributed Deep Learning:
  A Survey}.
\newblock \emph{arXiv e-prints}, art. arXiv:2003.03009, March 2020.

\bibitem[{Povey} et~al.(2014){Povey}, {Zhang}, and {Khudanpur}]{PoveyParallel}
Daniel {Povey}, Xiaohui {Zhang}, and Sanjeev {Khudanpur}.
\newblock {Parallel training of DNNs with Natural Gradient and Parameter
  Averaging}.
\newblock \emph{arXiv e-prints}, art. arXiv:1410.7455, Oct 2014.

\bibitem[Recht and Ré(2011)]{PSG}
Benjamin Recht and Christopher Ré.
\newblock Parallel stochastic gradient algorithms for large-scale matrix
  completion.
\newblock \emph{Mathematical Programming Computation}, 5, 04 2011.
\newblock \doi{10.1007/s12532-013-0053-8}.

\bibitem[{Sattler} et~al.(2018){Sattler}, {Wiedemann}, {M{\"u}ller}, and
  {Samek}]{2018arXiv180508768S}
Felix {Sattler}, Simon {Wiedemann}, Klaus-Robert {M{\"u}ller}, and Wojciech
  {Samek}.
\newblock {Sparse Binary Compression: Towards Distributed Deep Learning with
  minimal Communication}.
\newblock \emph{arXiv e-prints}, art. arXiv:1805.08768, May 2018.

\bibitem[{Shi} et~al.(2019){Shi}, {Wang}, {Zhao}, {Tang}, {Wang}, {Huang}, and
  {Chu}]{ShiA}
Shaohuai {Shi}, Qiang {Wang}, Kaiyong {Zhao}, Zhenheng {Tang}, Yuxin {Wang},
  Xiang {Huang}, and Xiaowen {Chu}.
\newblock {A Distributed Synchronous SGD Algorithm with Global Top-$k$
  Sparsification for Low Bandwidth Networks}.
\newblock \emph{arXiv e-prints}, art. arXiv:1901.04359, Jan 2019.

\bibitem[{Stich}(2018)]{F-SGD}
Sebastian~U. {Stich}.
\newblock {Local SGD Converges Fast and Communicates Little}.
\newblock \emph{arXiv e-prints}, art. arXiv:1805.09767, May 2018.

\bibitem[{Su} and {Chen}(2015)]{EMA}
Hang {Su} and Haoyu {Chen}.
\newblock {Experiments on Parallel Training of Deep Neural Network using Model
  Averaging}.
\newblock \emph{arXiv e-prints}, art. arXiv:1507.01239, July 2015.

\bibitem[Wen et~al.(2017)Wen, Xu, Yan, Wu, Wang, Chen, and Li]{WenXu17Terngrad}
Wei Wen, Cong Xu, Feng Yan, Chunpeng Wu, Yandan Wang, Yiran Chen, and Hai Li.
\newblock Terngrad: Ternary gradients to reduce communication in distributed
  deep learning.
\newblock In \emph{Advances in neural information processing systems}, pages
  1509--1519, 2017.
\newblock URL \url{https://arxiv.org/abs/1705.07878}.

\bibitem[Yu et~al.(2019)Yu, Yang, and Zhu]{PRSGD}
Hao Yu, Sen Yang, and Shenghuo Zhu.
\newblock Parallel restarted sgd with faster convergence and less
  communication: Demystifying why model averaging works for deep learning.
\newblock \emph{Proceedings of the AAAI Conference on Artificial Intelligence},
  33:\penalty0 5693--5700, 07 2019.
\newblock \doi{10.1609/aaai.v33i01.33015693}.

\bibitem[{Zhang} et~al.(2016){Zhang}, {De Sa}, {Mitliagkas}, and
  {R{\'e}}]{Parallel-SGD}
Jian {Zhang}, Christopher {De Sa}, Ioannis {Mitliagkas}, and Christopher
  {R{\'e}}.
\newblock {Parallel SGD: When does averaging help?}
\newblock \emph{arXiv e-prints}, art. arXiv:1606.07365, June 2016.

\bibitem[Zhang et~al.(2018)Zhang, Tu, Ren, Wan, Zhou, Li, and Wang]{SYPS}
Jilin Zhang, Hangdi Tu, Yongjian Ren, Jian Wan, Li~Zhou, Mingwei Li, and Jue
  Wang.
\newblock An adaptive synchronous parallel strategy for distributed machine
  learning.
\newblock \emph{IEEE Access}, pages 1--1, 03 2018.
\newblock \doi{10.1109/ACCESS.2018.2820899}.

\bibitem[Zhang et~al.(2015)Zhang, Choromanska, and LeCun]{elastic}
Sixin Zhang, Anna~E Choromanska, and Yann LeCun.
\newblock Deep learning with elastic averaging sgd.
\newblock In C.~Cortes, N.~D. Lawrence, D.~D. Lee, M.~Sugiyama, and R.~Garnett,
  editors, \emph{Advances in Neural Information Processing Systems 28}, pages
  685--693. Curran Associates, Inc., 2015.
\newblock URL
  \url{http://papers.nips.cc/paper/5761-deep-learning-with-elastic-averaging-sgd.pdf}.

\bibitem[{Zhang} et~al.(2015){Zhang}, {Gupta}, {Lian}, and
  {Liu}]{2015Staleness-aware}
Wei {Zhang}, Suyog {Gupta}, Xiangru {Lian}, and Ji~{Liu}.
\newblock {Staleness-aware Async-SGD for Distributed Deep Learning}.
\newblock \emph{arXiv e-prints}, art. arXiv:1511.05950, Nov 2015.

\bibitem[{Zhao} and {Canny}(2014)]{zhao-Kylix}
H.~{Zhao} and J.~{Canny}.
\newblock Kylix: A sparse allreduce for commodity clusters.
\newblock In \emph{2014 43rd International Conference on Parallel Processing},
  pages 273--282, Sep. 2014.
\newblock \doi{10.1109/ICPP.2014.36}.

\bibitem[Zheng et~al.(2017)Zheng, Meng, Wang, Chen, Yu, Ma, and
  Liu]{ASGDwithDC}
Shuxin Zheng, Qi~Meng, Taifeng Wang, Wei Chen, Nenghai Yu, Zhi-Ming Ma, and
  Tie-Yan Liu.
\newblock Asynchronous stochastic gradient descent with delay compensation.
\newblock In \emph{Proceedings of the 34th International Conference on Machine
  Learning-Volume 70}, pages 4120--4129. JMLR. org, 2017.

\bibitem[Zinkevich et~al.(2010)Zinkevich, Weimer, Smola, and Li]{PSGD}
Martin Zinkevich, Markus Weimer, Alexander Smola, and Lihong Li.
\newblock Parallelized stochastic gradient descent.
\newblock volume~23, pages 2595--2603, 01 2010.

\end{thebibliography}
\newpage
\appendix
    \section{The $DynamicDatasetAdjust$ algorithm}
    \label{alg:da}
    The workflow of the $DynamicDatasetAdjust$ function is shown in Algorithm \ref{alg:datasetAdjust}. The batch size $B_i^{'j+1}$ of the worker $i$ is assumed to be rounded down at step \ref{dds:step1}. In order to keep the variance of rounding minimum, steps from \ref{dds:step2} to \ref{dds:step3} pick no more than $k$ values and round them up. After rounding twice, the $B^{'j+1}$ is a set of the exact adjusted batch size for the next training epoch. Then, the $B^{'j+1}$ is normalized, and unitized to compute the range of the sub-dataset for each worker. Note that the calculation of $B^{'j+1}$ and $[L^{j+1}, K^{j+1}]$ will be repeated until the distributed DNN training reaches the end. 
    
    \begin{algorithm}[h] 
        \caption{DynamicDatasetAdjust} 
        \label{alg:datasetAdjust} 
        \begin{algorithmic}[1] 
            \Require 
            A list of adjusted batch sizes of workers: $[B_1^{j+1}, B_2^{j+1}, B_i^{j+1}, \cdots, B_n^{j+1}]$;
            \Ensure 
            The range of sub dataset for each worker, the size of the lists is equal to the size of the cluster: $(L^j, K^j)$;
            \State Initialization: $(L^j=[0.0], K^j=[0.0])$;
            \State $B_i^{'j+1} = roundingDown(B_i^{j+1})$;
            \label{dds:step1}
            \State $k = disparity(B_i^{'j+1}, B_i^{j+1})$;
            \State $[id_1, id_2, \cdots, id_k] = argsort(B_{decimal}^{j+1})$;
            \label{dds:step2}
            \State $B_i^{'j+1}=roudingUp(B_i^{'j+1}, k, B_{decimal}^{j+1})$
            \label{dds:step3}
            \State $Norm(B^{('j+1)})$;
            \For {$i=1, n$;}
            \State $L_i^{j+1} = \sum_{i=0}^{i-1} B_{i}^{('j+1)}$;
            \State $K_i^{j+1} = L_i^{j+1} + B_{i}^{('j+1)}$;
            \EndFor
        \\\Return $(L^{j+1}, K^{j+1})$;
        \end{algorithmic} 
    \end{algorithm}
         For example, assuming total batch size $B=64$, $B^{j+1}=[13.7, 16.5, 19.6, 14.2]$. After rounding down, $B^{'j+1}=[13, 16, 19, 14]$, so the $k = B - (13 + 19 + 16 + 14) = 2$. Then $B^{j+1}$ is sorted by its decimal fraction, because $0.7 \geq 0.6 \geq 0.5 \geq 0.3$, the indexes of top-$k$ values is $[0, 2]$. After rounding up, the $B^{'j+1}$ is updated to $[14,16,20,14]$. After the normalization of the $B^{'j+1}$,  the ratio of the $B^{'j+1}$ is [0.22,~0.25,~0.31,~0.22], and the dynamic partition $(L^{j+1}, K^{j+1})$ is [[0,~0.22],~[0.22,~0.47],~[0.47,~0.78],~[0.78,~1]].
   \section{More experiments}
    \begin{figure}[!h]
        \begin{subfigure}{0.48\textwidth}            \includegraphics[width=\linewidth]{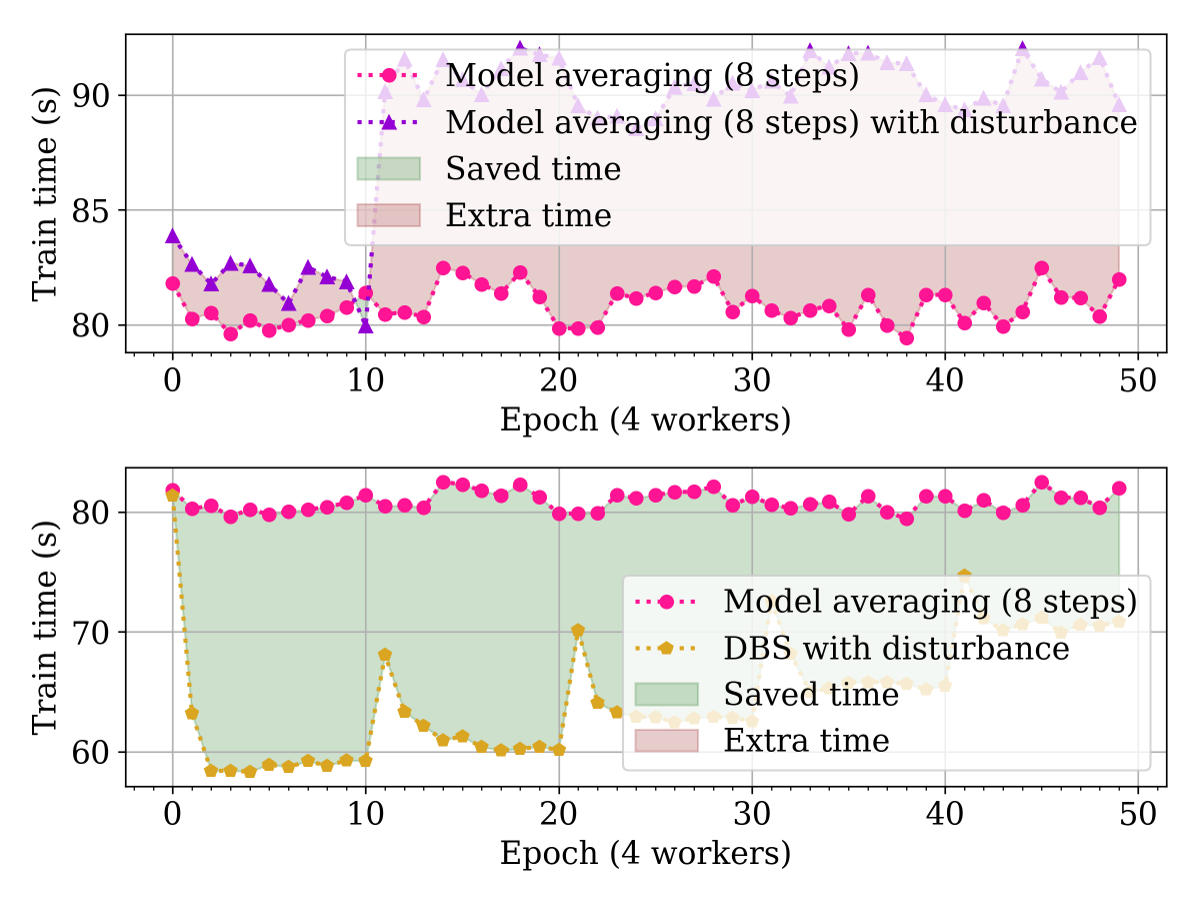}
            \caption{The robustness comparison of the S-SGD and the model averaging method with disturbance. For example, the synchronous period is 8 iterations~ (i.e., the synchronization interval:$step = 8$).} \label{fig:model-averaging}
        \end{subfigure}%
        \hspace*{\fill}   % maximize separation between the subfigures
        \begin{subfigure}{0.48\textwidth}
            \includegraphics[width=\linewidth]{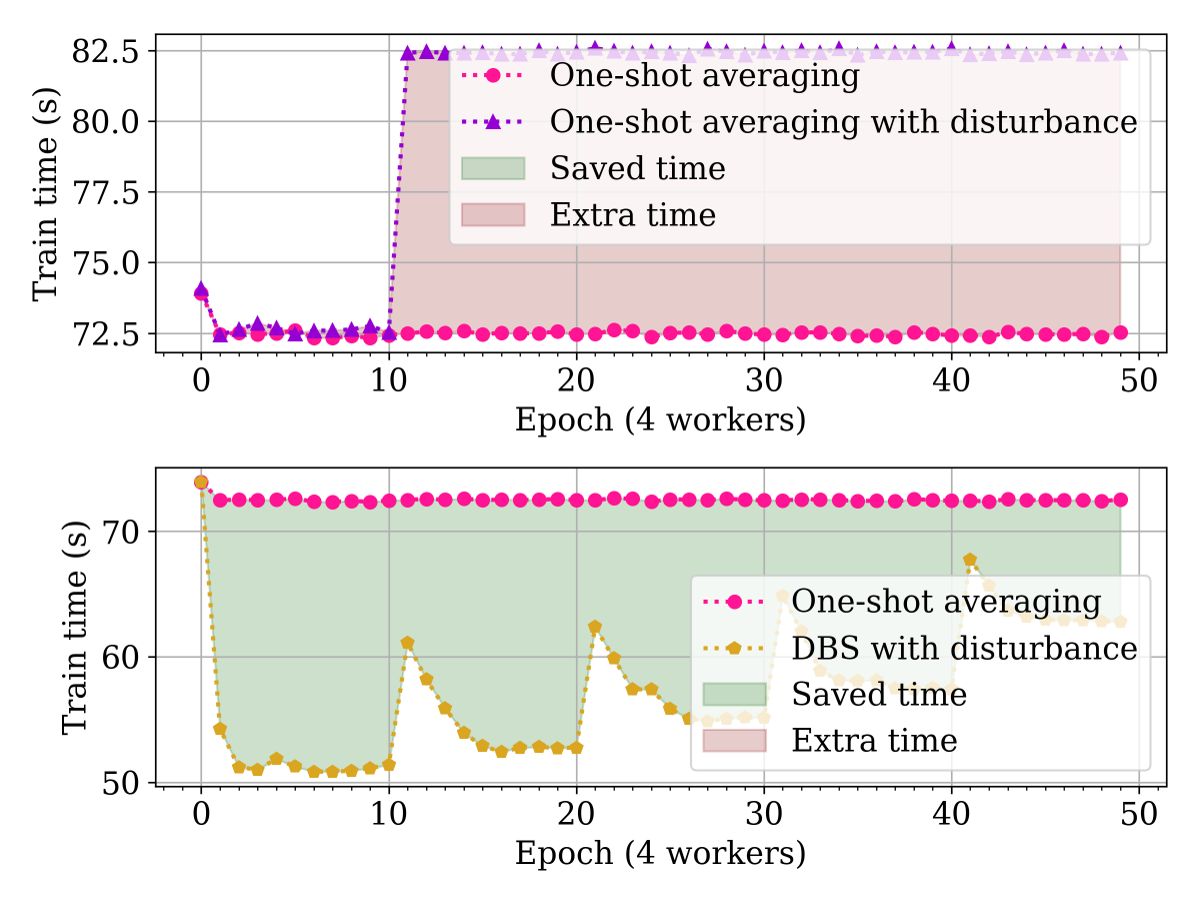}
            \caption{The robustness comparison of the S-SGD and the one-shot method with disturbance.} \label{fig:one-shot}
        \end{subfigure}%
        \label{fig:model averaging}
        \caption{The the consumed time of the model averaging and one-shot with fixed batch size increases rapidly because of the disturbance, while the DBS strategy can still save lots of time.}
    \end{figure}

   \section{Lemmas}
   \newtheorem{lemma}{Lemma}
   \begin{lemma}    \label{lm1}
   Under the assumption in Theorem \ref{ta:t1}, and m the mini-batch size, larger the m, lower the  $\mathbb{D}[ \frac{\sum_{i=1}^{m} f(x_{i})}{m} ]$ we have.\\
   \end{lemma}
   
   \newtheorem{proof}{Proof}
   \begin{proof}	\label{prf:lm1}
   let $C = \mathop{max}\{Cov[f(x_{i}),f(x_{k})]\}$, $i \neq k$
   		\begin{equation}
   		\begin{aligned}
   		\mathbb{D}[ \frac{\sum_{1}^{m} f(x_{i})}{m} ] & = \frac{\mathbb{D}[ \sum_{i=1}^{m} f(x_{i})]}{m^{2}} \\
   		& = \frac{1}{m^{2}}\{m\mathbb{D}{f(x_{i})} + \sum_{i=1}^{m}\sum_{k \neq i}Cov[f(x_{i}),f(x_{k})]\} \\
   		& \leq \frac{1}{m^{2}}\{m\mathbb{D}{f(x_{i})} + m(m-1)C \} \\
   		& = \frac{1}{m}[\mathbb{D}{f(x_{i})} - C] + C
   		\end{aligned}
   		\end{equation}
   \end{proof}
	
    \section{Proof of Theorem \ref{ta:t1}}
    \begin{proof}
    \label{prf:ta:t1}
        let $d^{j} = x^{j} - x^{*}$
        \begin{equation}
        \begin{aligned}
        \| d^{j+1} \|^{2} & = \| d^{j }- \gamma\nabla f_{i}(x^{j}) \|^{2}\\
        & = \| d^{j} \| ^{2} - 2\gamma\langle d^{j}, \nabla f_{i}(x^{j})\rangle + \gamma^{2} \| \nabla f_{i}(x^{j}) \|^{2}\\
        \end{aligned}
        \label{ta:8}
        \end{equation}
        by (\ref{ta:2}) and (\ref{ta:6}), Taking expectation conditioned on $x^{j}$:
        \begin{equation}
        \begin{aligned}
        \mathbb{E}\|d^{j+1}\| ^2 & \le (1-\gamma\mu) \mathbb{E} \| d^{j} \| ^{2} - 2\gamma(f(x)-f(x^{*})) + \gamma ^{2} \sigma ^{2}\\
        & \le  (1-\gamma\mu) \mathbb{E} \| d^{j} \| ^{2} + \gamma ^{2} \sigma ^{2}
        \end{aligned}
        \label{ta:9}
        \end{equation}
        by solving the recursive inequality:
        \begin{equation}
        \begin{aligned}
        \mathbb{E} \| d^{j} \| ^{2} & \le (1-\gamma\mu)^{j} \mathbb{E} \| d^{0} \| ^{2} + \sum_{i=1}^{j} \gamma ^{2} \sigma ^{2} (1-\gamma\mu)^{i-1}\\
        & \le (1-\gamma\mu)^{j} \| x^{0} - x^{*} \| ^{2} + \frac{\gamma\sigma^{2}}{\mu}
        \end{aligned}
        \label{ta:10}
        \end{equation}
        where we use geometric series summing-up in the last step.\\
    \end{proof}
    
\end{document}